\documentclass{ieeeaccess}
\pdfoutput=1
\usepackage{cite}
\usepackage{amsmath,amssymb,amsfonts}
\usepackage{algorithmic}
\usepackage{graphicx}
\usepackage{textcomp}
\usepackage{caption}
\usepackage{booktabs}
\usepackage{multirow}
\usepackage{url}
\usepackage{array}
\newcommand{\PreserveBackslash}[1]{\let\temp=\\#1\let\\=\temp}
\newcolumntype{C}[1]{>{\PreserveBackslash\centering}p{#1}}
\newcolumntype{L}[1]{>{\PreserveBackslash\raggedright}p{#1}}
\def\BibTeX{{\rm B\kern-.05em{\sc i\kern-.025em b}\kern-.08em
    T\kern-.1667em\lower.7ex\hbox{E}\kern-.125emX}}
\begin{document}
\history{Date of publication xxxx 00, 0000, date of current version xxxx 00, 0000.}
\doi{10.1109/ACCESS.2020.DOI}

\title{Event Arguments Extraction via Dilate Gated Convolutional Neural Network with Enhanced Local Features}
\author{\uppercase{Zhigang Kan}\authorrefmark{1},
\uppercase{Linbo Qiao\authorrefmark{2}, Sen Yang\authorrefmark{3}, Feng Liu\authorrefmark{4}, and Feng Huang
}\authorrefmark{5}}
\address[1]{School of Computer, National University of Defense Technology, Changsha, China
(e-mail: kanzhigang5206@163.com)}
\address[2]{School of Computer, National University of Defense Technology, Changsha, China
(e-mail: qiao.linbo@nudt.edu.cn)}
\address[3]{School of Computer, National University of Defense Technology, Changsha, China
(e-mail: yangsen.nudt@hotmail.com)}
\address[4]{School of Computer, National University of Defense Technology, Changsha, China
(e-mail: richardlf@nudt.edu.cn)}
\address[5]{School of Computer, National University of Defense Technology, Changsha, China
(e-mail: fhuang@nudt.edu.cn)}

\tfootnote{Project supported by the National Key R\&D Program of China (No. 2018YFB0204301) and the National Natural Science Foundation of China (No. 61806216)}

\markboth
{Z.Kan \headeretal: Event Arguments Extraction via Dilate Gated Convolutional Neural Network with Enhanced Local Features}
{Z.Kan \headeretal: Event Arguments Extraction via Dilate Gated Convolutional Neural Network with Enhanced Local Features}

\corresp{Corresponding author: Linbo Qiao (e-mail: qiao.linbo@nudt.edu.cn).}

\begin{abstract}
Event Extraction plays an important role in information-extraction to understand the world.
Event extraction could be split into two subtasks: one is event trigger extraction, the other is event arguments extraction.
However, the F-Score of event arguments extraction is much lower than that of event trigger extraction, i.e. in the most recent work, event trigger extraction achieves 80.7\%, while event arguments extraction achieves only 58\%.
In pipelined structures, the difficulty of event arguments extraction lies in its lack of classification feature, and the much higher computation consumption.
In this work, we proposed a novel Event Extraction approach based on multi-layer Dilate Gated Convolutional Neural Network (EE-DGCNN) which has fewer parameters.
In addition, enhanced local information is incorporated into word features, to assign event arguments roles for triggers predicted by the first subtask.
The numerical experiments demonstrated significant performance improvement beyond state-of-art event extraction approaches on real-world datasets.
Further analysis of extraction procedure is presented, as well as experiments are conducted to analyze impact factors related to the performance improvement.
\end{abstract}


\titlepgskip=-15pt

\maketitle

\section{Introduction}
\label{sec:introduction}
\PARstart{T}{he} global web with electronic information, including most notably the WWW, provides a huge resource of unbounded information to understand the world.
Most text data from social media and the Internet is unstructured in the machine view and merely composed by some human-readable natural languages.
But to fully exploit the information requires the ability to extract unstructured content automatically.
To get interesting, representative and human-interpretable information from those text data, many data mining techniques have been created, and various kinds of algorithms implemented to develop the capability to extract structured information from data sources.
Among these techniques, one of the most promising way to obtain hidden knowledge is Event Extraction, which was defined in Automatic Content Extraction (ACE)\cite{Doddington2004TheAC} evaluation.
An event is a specific occurrence involving participants, it is something that happens, it can frequently be described as a change of state.

\begin{figure*}[th]
	\centering
    \includegraphics[scale=1]{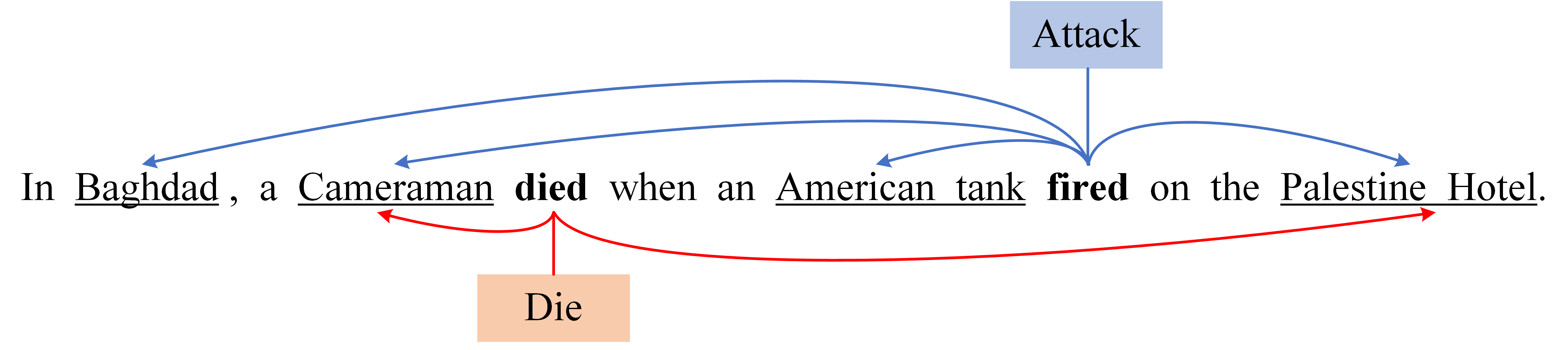}
	\caption{An example for event extraction task. There are two events in this event mention: "Attack" and "Die". Bold words are triggers while words with underline are event arguments.}
	\label{EEdemo}
\end{figure*}

Event extraction becomes a useful technique obtained more and more attention in the research community and industry applications for information retrieval and information extraction systems.
It has been studied for a long time, both from the perspective of philosophy and the perspective of machine learning, but still remains a great challenging task.
The purpose of event extraction is to determine the type of events and extract arguments with different roles from unstructured text automatically by machine.
With the development of knowledge graph, good event type and argument roles extraction are expected to significantly help information retrieval and information extraction systems.

Considering that, natural language exhibits syntactic properties that would naturally combine words to phrases.
Order-insensitive models are insufficient to fully capture the semantics of natural language due to their inability to account for differences in meaning as a result of differences in word order or syntactic structure.
We, therefore, apply contextual sensitive sequential models to extract event triggers and arguments.
In particular, bidirectional transformer models are a linguistically attractive option due to their relation to syntactic interpretations of sentence structure, which is promising in event extraction.

Due to the complexity of the event extraction task, researchers usually build complex neural networks to train the classification model end-to-end by force. The performances of these neural network models are good, but the numbers of their parameters are large. This results in prohibitively computational consumption.

In this paper, to improve the accuracy of assigning argument roles, we propose an elegant framework which has fewer parameters to model relationships among context in an order sensitive setting, as well as using the pre-trained representations with local features
(The representations are obtained from large-scale open domain textual corpus, and the local features are obtained from the input).
Specifically, our work makes the following contributions:
\begin{itemize}
\itemsep -1pt
\item A multi-layer dilate gated convolutional neural network which has an advantage in the number of parameters is constructed elaborately to extract event arguments via reformulating the event extraction problem as a sequential tagging problem.
\item Furthermore, local features among event trigger and event arguments in an event mention are added to the proposed Dilate Gated Convolutional Neural Network based Event Extraction (EE-DGCNN) model input representation, which significantly improved the identification accuracy and classification accuracy of the extraction model.
\item  A serial of numerical experiments are conducted and the experimental results demonstrate the efficacy of the proposed EE-DGCNN in event extraction task. Furthermore, the analysis of performance related factors on the event extraction is presented. As well as comparative experiments conducted to choose an event extraction system with better performance.
\end{itemize}

\section{Related Works}

\subsection{Event Extraction}

Event extraction task in this paper was defined in Automatic Content Extraction (ACE) evaluation\footnote{http://projects.ldc.upenn.edu/ace/}.
ACE defines an event as a specific occurrence involving participants, which frequently expressed as something that happens or a change of state.
We introduce several ACE terminology used in this paper as follows:
\begin{itemize}
\itemsep -1pt
\item \textbf{Event mention:} a sentence or phrase within which an event is described.
\item \textbf{Event trigger:} the word or phrase that most clearly expresses the occurrence of an event.
\item \textbf{Event argument:} the participants or attributes of an event. They are mainly entity mentions, temporal expressions or values (e.g. \textit{Job-Title}).
\item \textbf{Argument role:} the role an argument played in the event which it participates in.
\end{itemize}

The goal of the event extraction task is to extract structured event information from text that contains event information.
Specifically, when processing the event mention in Fig. \ref{EEdemo}, researchers need to find the event triggers ("fired" and "died") to determine the event type("Attack" and "Die") and extract event arguments for them. Table \ref{Structured_event_information} shows the structured event information extracted from the event mention in Fig. \ref{EEdemo}.

\begin{table}[thp]
	\centering
	\caption{Structured event information from the event mention in Fig. \ref{EEdemo}.}
	\addtolength{\tabcolsep}{4.8pt}
	\begin{tabular}{|c|c|c|c|}
		\hline
        \multicolumn{4}{|c|}{Event 1} \\
        \hline
		Trigger & Event type & Argument & Role\\
		\hline
        \multirow{2}{*}{died} & \multirow{2}{*}{Die} & Cameraman & Victim \\

        \cline{3-4}
        ~ & ~ & Palestine Hotel & Place \\
		\hline
        \multicolumn{4}{|c|}{Event 2} \\
        \hline
		Trigger & Event type & Argument & Role\\
		\hline
        \multirow{4}{*}{fired} & \multirow{4}{*}{Attack} & Baghdad & Place \\

        \cline{3-4}
        ~ & ~ & Cameraman & Target \\
        \cline{3-4}
        ~ & ~ & American tank & Attacker \\
        \cline{3-4}
        ~ & ~ & Palestine Hotel & Place \\
		\hline
	\end{tabular}
	\label{Structured_event_information}
\end{table}

Considering the difference of the model structures, event extraction approaches could be categorized into two classes: one is pipeline structured, and the other is joint structured.
Early event extraction work mainly used pipeline structure with two stages, in which event trigger prediction applied at the first stage, and corresponding arguments classification conducted at the second stage.
Traditionally, multi-class classifiers are adopted in these stages to reduce the complexity of task \cite{Ahn2006stageseventextraction}.
Along this paradigm, many researchers abstracted higher-level features rather than beyond citizen-level to improve extraction performance \cite{Ji2008Refiningeventextraction,Hong2011Usingcrossentity,Hogenboom2011overvieweventextraction,Ritter2012Opendomainevent,Huang2016Liberaleventextraction}.

To avoid the inherent error propagation of pipeline approaches, researchers proposed joint extraction models.
Typical representatives of these methods are Markov Logic \cite{Poon2010Jointinferenceknowledge},
structured perceptron \cite{Li2013Jointeventextraction},
and dependency parsing \cite{McClosky2011Eventextractionas}.

In recent years, neural networks were used for event extraction tasks. DMCNN \cite{Chen2015Eventextractionvia} exploited dynamic multi-Pooling CNNs on the basis of the pipeline model while Nguyen et al. \cite{Nguyen2016Jointeventextraction} utilized RNNs to perform joint event extraction.
Both of them have achieved considerable performance, and suffer from a huge number of model parameters.
These neural network-based methods usually combine the input word vector with a variety of sentence-level and word-level information and apply a carefully constructed neural network to capture word features.
After the pre-trained language model appeared, its ability to express semantic information attracted the attention of researchers in the field of event extraction.
The PLMEE \cite{yang-plmee} method proposed by Yang et al. applied a pre-trained language model as a method to capture word features directly and obtained a large performance improvement.

\subsection{Word Representation}
Contextual features plays a very important role in event extraction.
Traditional word embedding, i.e. word2vec \cite{Mikolov2013Efficientestimationword}, methods apply a fixed pre-trained word embedding to represent the same word in different sentences, even if the word is polysemous.
Obviously, such a method is flawed and ignores the features of the context.

In order to utilize contextual information for word representation, ELMo (Embeddings from Language Models) was proposed \cite{Matthew2018ELMo} to capture features of the context with double layer bidirectional LSTM, which consists of a forward and backward language model.
Compared with the traditional word embedding model, word representation by ELMo has greatly improved the performance of applications.

Besides, a concise and explicit network architecture named transformer \cite{Vaswani2017transformer} which based on attention mechanisms were proposed to serve for machine translation tasks in the year 2017.
Benefiting from the attention mechanism, the transformer combines each word with a distant context when processing all the words in the sequence in parallel.
Taking into account this advantage of the transformer, the OpenAI team used a left-to-right transformer network instead of LSTM as a language model to capture long-distance language structures and proposed OpenAI GPT \cite{Radford2018Improvinglanguageunderstanding}.

Considering that the OpenAI GPT model employs a left-to-right transformer and the ELMo model simply splice embeddings at the highest level of two unidirectional LSTMs, they are not a true bidirectional language model.
Therefore, a model named BERT \cite{BERT2018}, which is based on bidirectional transformer, was proposed by Google AI Language team.
The BERT model has refreshed records on 11 NLP tasks, which shows that it is a very effective word representation model. Apart from the model proposed, training corpus is also plays an important role
\cite{Liu2016ExternalResources,Chen2017DistantSupervision}.

\begin{figure*}[tbh]
	\centering
    \includegraphics[scale=1.3]{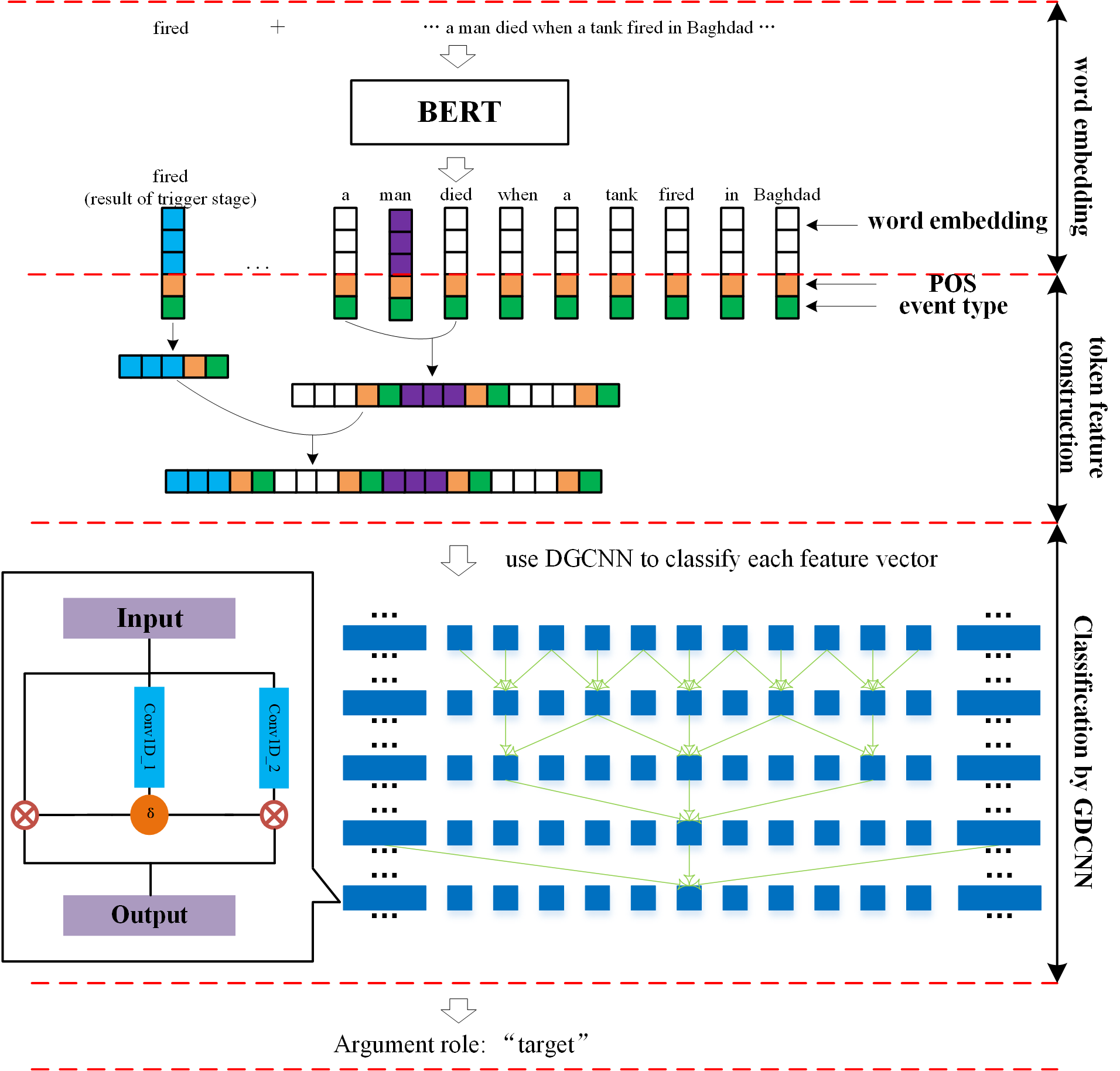}
	\caption{Proposed model for argument extraction stage. This is the second stage, and the first stage is to extract event trigger. After the first stage, the event trigger "fired" is identified and classified to be attack event type. A multi-layers dilate gated convolutional neural network is used to extract the token "man" to be the event argument corresponding to event trigger "fired".}
	\label{fine_tune_model_fig}
\end{figure*}

\section{Methodology}

We decompose the event extraction into two subtasks: {\it trigger classification} and {\it argument classification}.
In the first subtask, we identify trigger words by classifying each word in the sentence.
In the second subtask, for the triggers predicted in the first stage, we assign argument roles to each word in the sentence.

According to the specific details of the above two subtasks, we can consider the event extraction as tasks of sequence labeling.
Since the Bidirectional Transformer model owns the ability to capture contextual information, which is significant for extracting mentions, we propose to complete those two sequence labeling task with representations generated by BERT.
Thanks to the powerful language expression capabilities of the BERT model, a few layers of simple fully connected networks can achieve good results in the trigger extraction subtask.
We follow previous research work \cite{yang-plmee} to obtain triggers, and then assign argument roles via EE-DGCNN proposed in this paper.

As shown in Fig. \ref{fine_tune_model_fig}, the architecture of argument classification involves three basic components as follow:
\begin{itemize}
\itemsep -1pt
\item Word embeddings from a bidirectional transformer.
\item Constructing token features representation with lexical features and contextual features.
\item Assigning argument roles to each word in the sentence, with the predicted trigger.
\end{itemize}

\subsection{Fine-tuning via Bidirectional Transformer}

Generally speaking, there are three classes of neural networks applied to solve problems in natural language processing.
First ones are Recurrent Neural Networks (RNNs), which could be generally formulated as $y_t=f(y_{t-m}, \cdots, y_{t-1},x_t)$,
where $x_t$ is the current input variable, $f$ is a function with parameters to learn, and $y_{t-m}, \cdots, y_t$ are output.
To obtain a better capture of natural language, a bidirectional RNN is commonly suggested.
Second ones are Convolutional Neural Networks (CNNs), which could be basically formulated as $y_t = f(x_{t-m},\cdots, x_t, \cdots, x_{t+m})$,
in which the window size is $2m+1$.
The window size is highly related to capture the distribution of natural language.
And the third ones are Attention Mechanism (AM) based neural works \cite{Liu2017ArgumentAttention,Chen2018HBTNGMA}, which could be simply formulated as $y_t = f(x_t, A, B)$,
where $A, B$ are an external matrix to obtain attention.
Specifically, the function $f$ is defined as $f(Q,K,V)=\textnormal{softmax}\left( \frac{QK^\top}{\sqrt{d_k}}\right)V$.
The two most commonly used attention functions are additive attention, and dot-product (multiplicative) attention.
A lot of research works show that a better performance achieved with the attention mechanism adopted.

The purpose of the argument classification is to assign argument roles to words in the sentence where a predicted trigger is obtained.
Since the total number of argument roles defined by ACE is 36 (include "None"), the process of this subsection is to classify words into 36 categories using a multi-classifier.
We abstract the argument classification into a sequence labeling task: the input of argument classification is a sequence of words $(w_1, w_2, \cdots, w_n)$, and after processing, the model returns a sequence of labels $(l_1, l_2, \cdots, 1_n)$ which corresponding to the input words.
We obtain word embedding by applying a bidirectional transformer to pre-train the input word sequence and combine it with other features of the word to synthesize the token feature sequence. Details on the token features will be introduced in Section \ref{sec:tokenrep}.

In this work, we use the BERT, which is a language representation model based on a bidirectional transformer, to obtain pre-trained word embeddings.
This model pre-train the word sequence by using two unsupervised prediction tasks: Masked Language Model (MLM) and Next Sentence Prediction (NSP).
The MLM task randomly masks some percentage of the tokens from the input sequence, and the goal is to predict the masked words based only on its context.
The goal of the NSP task is to understand the relationship between the pair of sentences in the sequence of input.
Specifically, in this work, the NSP task is aim to capture the relationship between the predicted trigger and the original sentence.

We combine the word embeddings returned by the BERT with the other features of words and feed them into a multi-classifier to get the final label sequence.
The objective loss function of the multi-classifier is cross-entropy loss function.
The cross-entropy error function over a batch of multiple samples of size $n$ can be calculated as:

\begin{equation}
l(T, Y) = \sum_{i=1}^n l(t_i, y_i) = - \sum_{i=1}^n \sum_{c=1}^C t_{ic} \cdot \log(y_{ic})
\end{equation}

where $t_{ic}$ is 1 if and only if sample $i$ belongs to class $c$, and $y_{ic}$ is the output probability that sample $i$ belongs to class  $c$.
The probability that sample $i$ belongs to class $c$ is calculated by the softmax function $
y_{ic}=P(y=c|x)\frac{e^{x^\top w_c}}{\sum_{c=1}^C e^{x^\top w_c}}$,
where $w_c$ is the weight of the class $c$.

\subsection{Composition of Token Representation} \label{sec:tokenrep}

To make the bidirectional transformer more effective to subtasks, it is essential to add elaborately constructed features to word embeddings generated by the pre-trained model.
Specifically, in this work input representations of words in the sentence consists of the following three items:
\begin{itemize}
	\itemsep -1pt
	\item Output of pre-trained representation: The word embedding with the contextual information of the sentence generated by the pre-trained model. In this paper, it refers to the output of the BERT model.
	
	\item POS embeddings: embeddings for part-of-speech (POS) of tokens. Many words have multiple different parts of speech, we proposed to use POS embedding to alleviate this problem.
	
	\item Event type embeddings (Only in event argument classification subtask): Encodings for event type of current trigger. It's available for argument classification subtask. Since there may be multiple events in a sentence, we use event type embedding to enable the model to distinguish the event argument role of words in different events.
	Typical encoding strategies are included: (1) one-hot encoding; (2) using learnable vectors.
\end{itemize}

The potential relationship between the predicted trigger and the argument candidates is an important basis for argument classification. The input representation of the multi-classifier consists of the following parts:
\begin{itemize}
	\itemsep -1pt
	\item Trigger embedding: Trigger embedding is the feature of the predicted trigger which is the result of trigger classification subtask. It is concatenated by all the token embeddings of words in the current trigger.
	\item Candidate argument embedding: Context window refers to the current word and its context. When window size is 3, the context window contains the current word and a word to its left and right. Candidate argument embedding includes all token embeddings of the words in the context window.
\end{itemize}

\subsection{Multi-layer Dilate Gated Convolutional Neural Network}

Inspired by Wavenet architecture \cite{oord2016wavenet} which applied dilated convolution operations for audio synthesis, we enable the CNN model to capture farther distances without increasing model parameters, we used Dilate Convolutional Neural Network (DCNN).

\begin{figure}[ht]
	\centering
	\includegraphics[scale=0.9]{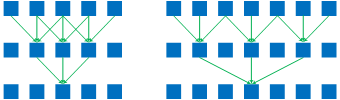}
	\caption{The left subfigure is the traditional Convolutional Neural Network, and the right subfigure is the Dilated Convolutional Neural Network.}
	\label{fig-dcnn}
\end{figure}

Considering a 3-layer convolutional neural network with a window size of 3 and the first layer is the input layer, which is shown vividly in Fig. \ref{fig-dcnn}.
If a traditional convolutional neural network is adopted, then each node can only capture 2 inputs before and after, and it is completely independent to other inputs.
However, if a dilate convolutional neural network is adopted, it can capture 3 inputs before and after, but the parameter quantity and speed do not change. This is because dilate convolutional node in the second layer skips the input directly adjacent to the center, directly capturing the center and the next adjacent input. In the third layer, three inputs are skipped in succession, and can also be regarded as a convolutional node with a window size of 7, which is related to the 7 original inputs.

\begin{figure}[ht]
	\centering
	\includegraphics[scale=1.8]{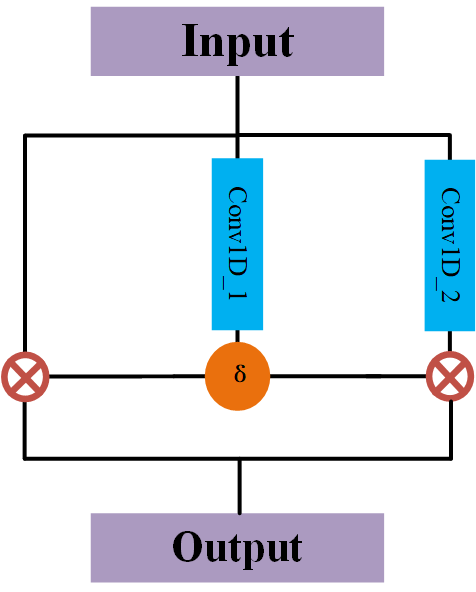}
	\caption{Gated Convolutional Block.}
	\label{fig-GCB}
\end{figure}

\begin{table*}[thp]
\centering
\caption{Overall performance on blind test data.}
\label{Comparison_to_state_of_the_art_methods}
\addtolength{\tabcolsep}{1pt}
\begin{tabular}{L{3cm}|C{1.6cm}|C{1.6cm}|C{1.6cm}|C{1.6cm}|C{1.6cm}|C{1.6cm}}
	\toprule[0.75pt]
	
	\multirow{1}*{Model} & \multicolumn{3}{c}{Argument Identification(\%)} & \multicolumn{3}{c}{Argument Role(\%)}  \\
	
	
	\midrule[0.5pt]
	~  & P &  R  & F & P  & R &  F\\
	
	\midrule[0.5pt]
	Li's baseline & 74.1 &37.4& 49.7 & 65.4 &33.1& 43.9 \\
	Lis structure & 69.8 &47.9& 56.8 & 64.7 &44.4&52.7\\
	Liao's cross-event & 50.9& 49.7& 50.3 & 45.1& 44.1 &44.6 \\
	Hongs cross-entity & 53.4 &52.9& 53.1 & 51.6 &45.5& 48.3\\
	Chen's DMCNN  & 68.8& 51.9& 59.1 & 62.2& 46.9& 53.5 \\
	Nguyen's JRNN & 61.4 &64.2 &62.8 & 54.2& 56.7& 55.4 \\
    Yang's PLMEE  & 71.4 &60.1 &65.3 & 62.3& 54.2& 58.0 \\
	\midrule[0.5pt]
	EE-DGCNN  & 71.7&61.3 &\textbf{66.1} &65.1 &57.7 & \textbf{61.2}  \\
	\bottomrule[0.75pt]
\end{tabular}
\end{table*}

And then, following the work of \cite{jianlin2019bdkgf}, a gated unit shown in Fig. \ref{fig-GCB} is added, which could be formulated as:
\begin{equation}
Y=Conv1D_2(X)\otimes \delta(Conv1D_1(X)),
\end{equation}
in which, two Conv1D is adopted in one gated unit. These two Conv1D is set to be with the same windows size, but the weights are independent to each other. The right Conv1D is activated by a function $\delta(\cdot)$. The output of the gated unit is the element-wise addition of these two Conv1Ds.


In the previous two subsections, we introduced the model for argument classification. This extraction framework can also be used to trigger subtask.
Compared to the argument classification, the input of the pre-training model could have only one sentence in the trigger classification subtask, and token embeddings are a combination of the output of pre-training and POS embeddings.

Pre-trained word embeddings are obtained by utilizing the BERT model.
We input a single sentence to the BERT model for trigger classification while a pair of sentences for argument classification.
It should be noted that, when dealing with sentence A, the sentence pair for argument classification is composed of sentence A and the trigger predicted at the first stage.

\section{Experiments}

\subsection{Details of experimental parameters}

\begin{figure}[htp]
	\centering
	\includegraphics[scale=1.55]{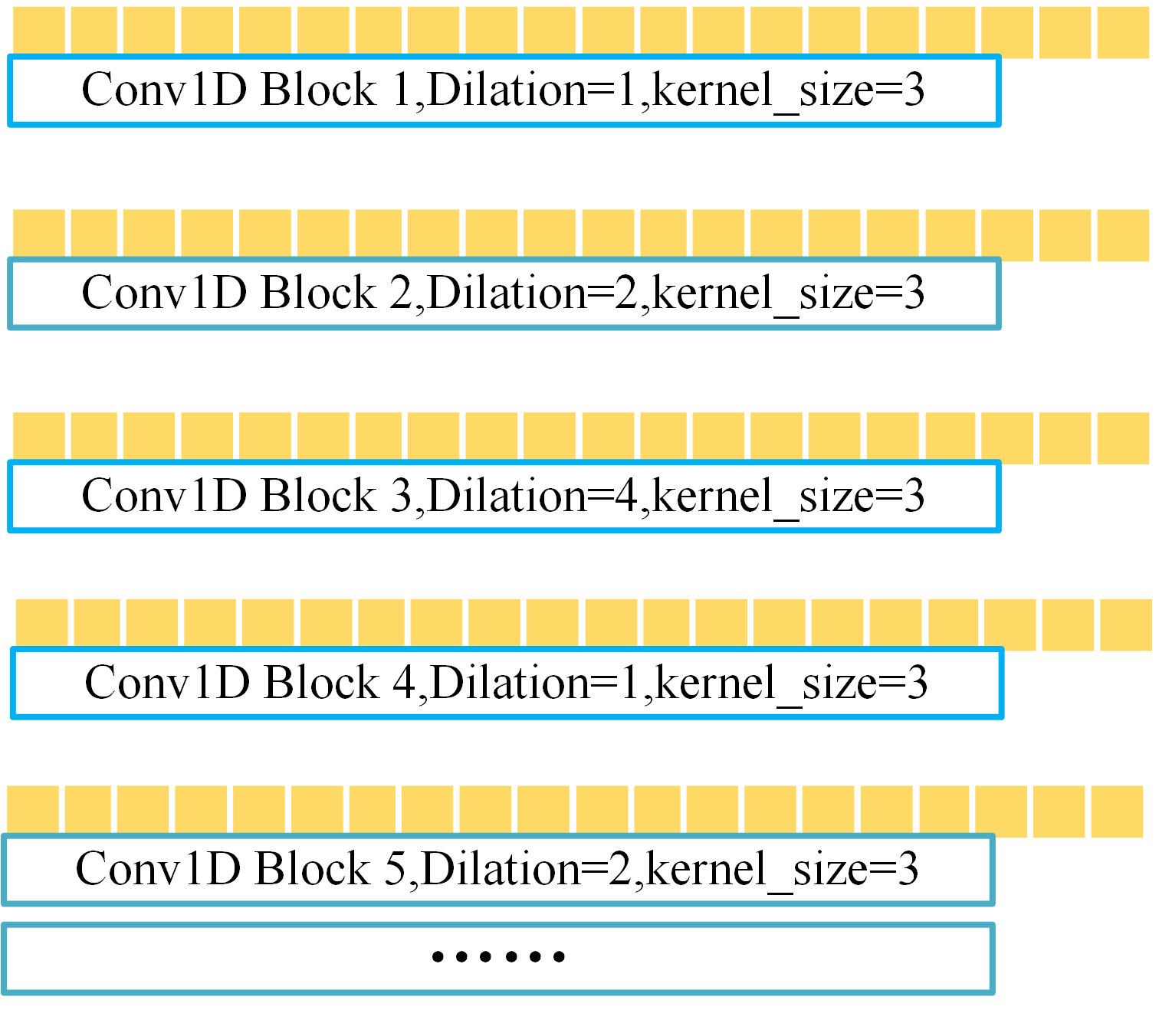}
	\caption{Gated Convolutional Neural Network with specified parameters in our experiments.}
	\label{fig-GCNN}
\end{figure}

Our fine-tuning procedure follows the framework proposed BERT.
We set bath size to 32, learning rate = $6e^{-5}$, the hidden dropout probability of the BERT model and the number of epochs is set to 0.2 and 40 respectively.
For the token feature, we use learnable vectors with 400 dimensions for event type embeddings and set window size to 3.
We use 7 layers of Conv1D in the multi-layer dilate gated convolutional neural network, and set the dilation to 1 every three layers. Fig. \ref{fig-GCNN} shows how the architecture of 7 layers of Conv1D.



\subsection{Dataset and Evaluation Metric}

We evaluate our model with the ACE 2005 corpus.
To verify the effectiveness of our EE-DGCNN model, we follow the data partitioning methods of previous research work \cite{Ji2008Refiningeventextraction,Liao2010Usingdocumentlevel,Li2013Jointeventextraction,Chen2015Eventextractionvia,Nguyen2016Jointeventextraction,yang-plmee}.
The test set of the experiment is composed of 40 newswire articles, while the validation set is 30 other documents from newswire. The remaining 529 articles of the ACE corpus make up the training set.

For the purpose of comparison, we follow the criteria of Chen \cite{Chen2015Eventextractionvia,yang-plmee} to determine the correctness of the predicted event mentions, and calculate the performance metric, namely P, R, and F1 score.

\subsection{Comparative Baselines}

State of the art algorithms and models for event extraction task are considered, and comparative baselines are listed as below:

\begin{itemize}
	\itemsep -1pt
	\item \textbf{Li's baseline \cite{Li2013Jointeventextraction}:}
	This is a pipelined method only uses local features.
	\item \textbf{Li's structure \cite{Li2013Jointeventextraction}:}
	This is the method that extracts events based on structure prediction.
	\item \textbf{Liao's cross-event \cite{Liao2010Usingdocumentlevel}:}
	The method in this work employs document-level information to improve performance.
	\item \textbf{Hong's cross-entity \cite{Hong2011Usingcrossentity}:}
	The method utilizes the cross-entity inference to extract event.
	\item \textbf{Chen's DMCNN model \cite{Chen2015Eventextractionvia}:}
	It is a pipelined system, which extracts event with dynamic multi-pooling convolutional neural networks.
	\item \textbf{Nguyen's JRNN model \cite{Nguyen2016Jointeventextraction}:}
	This is a joint system based on recurrent neural networks, which uses memory features to improve performance.
\item \textbf{Yang's PLMEE model \cite{yang-plmee}:}
	This is the best-reported pipelined system, which applies BERT as a pre-trained language model to capture word features.
To the best of our knowledge, it is the best-reported event extraction system based on pre-trained language model as well.
\end{itemize}

Table \ref{Comparison_to_state_of_the_art_methods} shows the best performance of our model and the methods mentioned above.
We can see that our EE-DGCNN model, which is a pipelined system, achieves the best performance among all the compared systems.
Compared to the best-reported model \cite{yang-plmee}, our model owns a great improvement over 3.2\% for argument role assignment.
This shows that our EE-DGCNN model which has richer local features and a better classifier is conducive to argument extraction.
In addition, EE-DGCNN surpasses JRNN which is a joint method over 5.8\% respectively.
The experimental results demonstrate the efficacy of the proposed model EE-DGCNN.

\subsection{Effect of Token Features}

We replace the token features fed to the multi-classifier only with the output of the BERT, and named it as EE-DGCNN-SimpleF.
To make these two methods comparable, we train them with the same parameters on the same data set.

\begin{table}[htp]
	\centering
	\caption{Comparison of F1 score for EE-DGCNN and EE-DGCNN-SimpleF models.}
	\addtolength{\tabcolsep}{4.8pt}
	\begin{tabular*}{7.9cm}{lcc}
		\toprule[0.75pt]
		System& Identification & Classification\\
		\midrule[0.5pt]
		SimpleF & 65.4\% & 60.5\% \\
		
		EE-DGCNN & {\textbf{66.1\%}} & {\textbf{61.2\%}} \\
		\bottomrule[0.75pt]
	\end{tabular*}
	\label{Comparison_EE-DGCNN_and_EE-DGCNN-SimpleF}
\end{table}

As shown in Table \ref{Comparison_EE-DGCNN_and_EE-DGCNN-SimpleF}, EE-DGCNN achieves an improvement of 0.7\% over EE-DGCNN-SimpleF on both Argument identification and argument role assignment.
The facts of those improvements confirm the effectiveness of the token feature.
It can be seen from the comparison, that although the BERT model has been trained on large scale open domain context, while event extraction with the word embedding generated by BERT still has much room for improvement.
As for event extraction task, the emphasis of grammar information about the word, such as POS, and information obtained in the first stage, such as event type for argument classification, enhance the model's ability to extract event in an event mention.

\subsection{Single Sentence vs. Sentence Pair for Argument Classification Input}

There are two options for the input of the pre-trained model in the argument classification phase: 1) single sentence; 2) sentence pair.
Specifically, when we deal with the sentence "A man died when a tank fired in Baghdad" and the predicted event trigger is "fired", these two inputs of BERT model are listed as follow:
\begin{itemize}
	\itemsep -1pt
	\item Single sentence (a sequence of tokens in the sentence):
	The predicted trigger "fired" was concatenated at the beginning of the sentence.
	Segment embeddings of tokens in this sequence are set to be 0.
	\item Sentence pair: The first sentence of this sentence pair is the obtained event trigger "fired", and the other is the sentence of event mention to deal with.
	The input is a token sequence composed of the sentence pair, and two sentences are separated by special characters.
	The segment embeddings for the first sentence is 0 while the second is 1.
\end{itemize}

\begin{table}[h]
	\centering
	\caption{Comparison of performance (F1 score) for argument classification base on different input representations.}
	\addtolength{\tabcolsep}{1.8pt}
	\begin{tabular*}{7.9cm}{lcc}
		\toprule[0.75pt]
		Input Form & Identification & Classification\\
		\midrule[0.5pt]
		Single sentence& 65.4\%&60.8\% \\
		
		Sentence pair& \textbf{66.1\%}&\textbf{61.2\%} \\
		\bottomrule[0.75pt]
	\end{tabular*}
	\label{Comparison_BERT_input}
\end{table}

Table \ref{Comparison_BERT_input} illustrates that the input of Sentence pair outperform the baseline with the input of single sentence.
Using sentence pairs as the input to the BERT brings a performance improvement to the model, because this input method may capture the potential relationship between triggers and arguments more effectively.

\subsection{Event Type Encoding Forms}

In the second stage, for event argument classification subtask, event type embeddings are very important features.
To investigate the appropriate method of encoding event types, we modify event type embeddings in EE-DGCNN to the following forms:
\begin{itemize}
	\item No Event Type: remove event type embeddings from token embeddings.
	\item One-hot codes: Use one-hot codes of event types as event type embeddings.
	\item 50D-vectors: Apply 50-dimensional learnable vectors to represent event type embeddings.
	\item 400D-vectors:  change the dimension of learnable vectors in 50D-vectors to 400.
\end{itemize}

\begin{table}[thp]
	\centering
	\caption{Performance (F1 score) of methods with different event type embeddings forms.}
	\addtolength{\tabcolsep}{1.8pt}
	\begin{tabular*}{7.9cm}{lcc}
		\toprule[0.75pt]
		Event type & \multirow{2}*{Identification} & \multirow{2}*{Classification}\\
		embeddings & ~ & ~\\
		\midrule[0.5pt]
		No Event Type&64.1\% &60.5\% \\
		
		One-hot codes&65.3\% &60.7\% \\
		
		50D-vectors&65.1\% &60.2\% \\
		
		400D-vectors&\textbf{66.1\%} &\textbf{61.2\%} \\
		\bottomrule[0.75pt]
	\end{tabular*}
	\label{Comparison_event_type_forms}
\end{table}

\begin{figure}[h]
	\centering
	\includegraphics[scale=0.35]{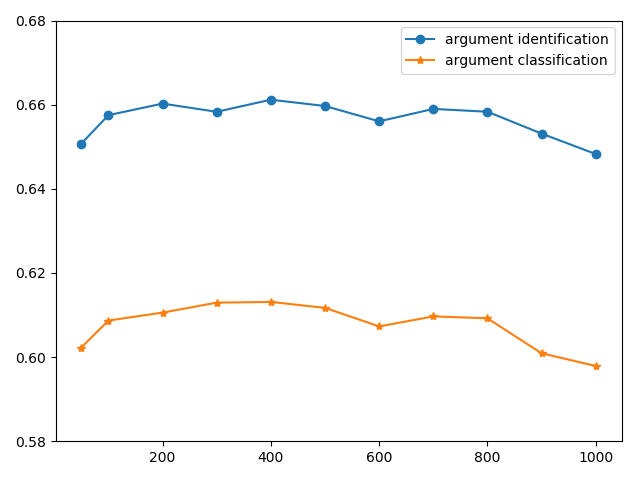}
	\caption{Performances for argument identification and classification with different dimensions of event type embeddings.}
	\label{argument_result}
\end{figure}

Table \ref{Comparison_event_type_forms} presents the performance (F1 scores) for argument identification and classification.
The method 400D-vectors achieves improvements of 0.5\% over the method One-hot codes and 0.7\% over No Event Type.
This demonstrates the benefit of learnable event type embeddings for the token feature in section \ref{sec:tokenrep}.
Further, in order to investigate the relationship between the dimensions of event type embeddings and its expressive ability, we modify the dimensions for more experiments.
Fig. \ref{argument_result} shows that EE-DGCNN achieves the best F1 score when the dimension set at 400.

\section{Conclusion}
In this paper, we propose a novel pipeline event extraction model based on a  dilate gated convolutional neural network, which utilizing word embedding generated by the pre-trained model, as well as elaborately constructed token features.
We designed an event extraction model based on the bidirectional transformer for event argument roles assignment at the second stage. Numerical experiments conducted on real-world dataset demonstrate the efficacy of the proposed model EE-DGCNN, and further analysis of the relevant factors related to extraction performance is conducted.

Based on the comparative analysis results, the proposed model is enhanced with elaborately selected strategies, which enjoy the performance beyond all the state of the art baselines with a significant improvement in F score.

\bibliographystyle{ieeetr}
\bibliography{access}

\begin{IEEEbiography}[{\includegraphics[width=1in,height=1.25in,clip,keepaspectratio]{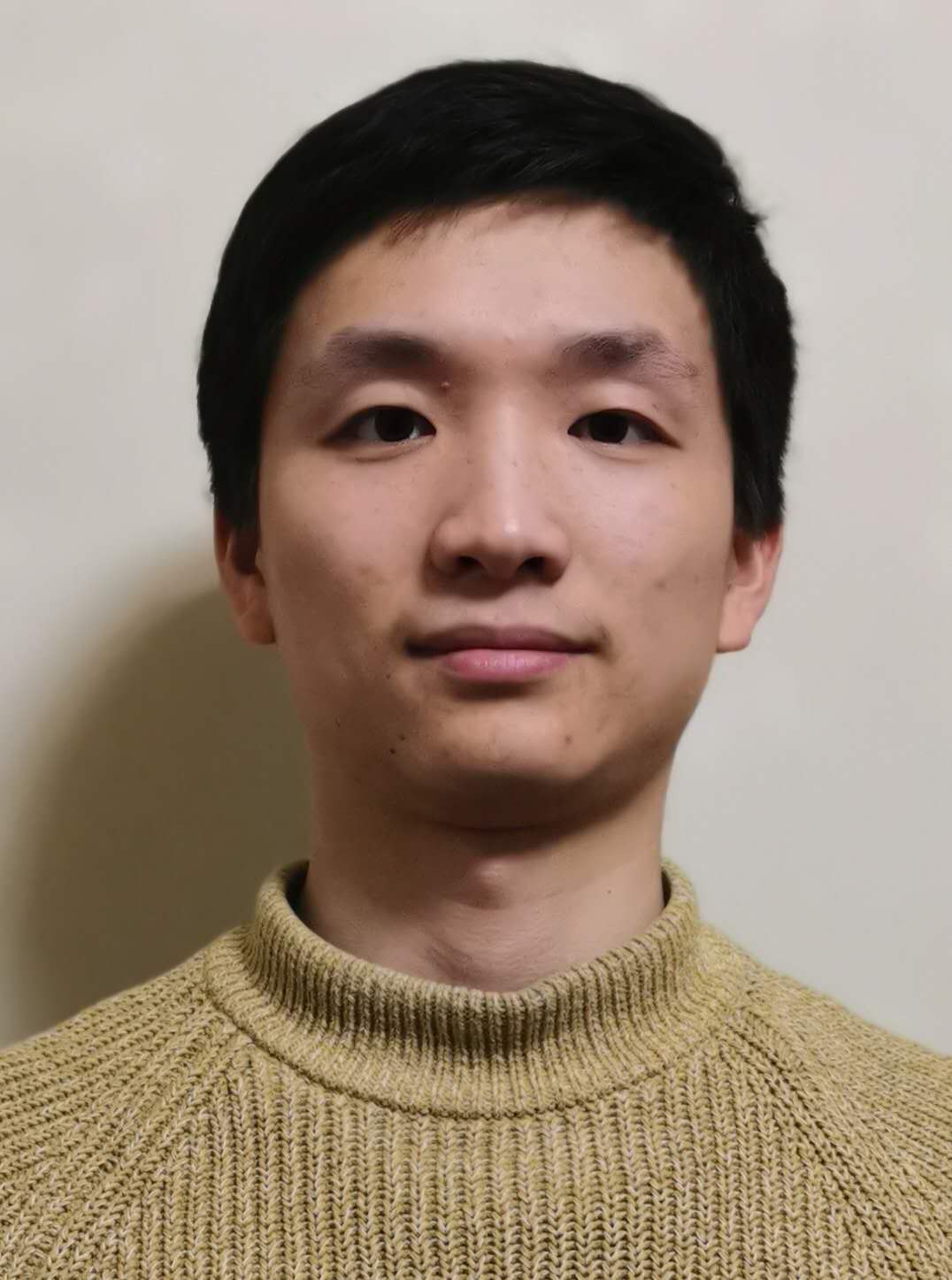}}]{Zhigang Kan} received his B.E. degree in computer science from the National University of Defense Technology (NUDT), Changsha, China, in 2017. His research interests include event extraction and natural language processing.
\end{IEEEbiography}

\begin{IEEEbiography}[{\includegraphics[width=1in,height=1.25in,clip,keepaspectratio]{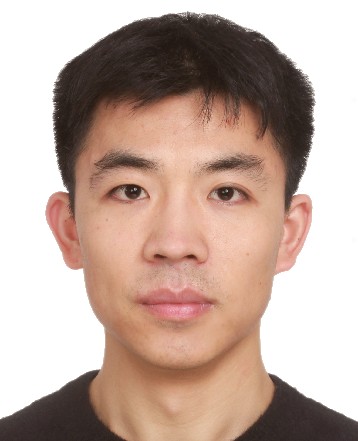}}]{Linbo Qiao} received the Ph.D., M.S., and B.S. degrees in computer science and technology from the National University of Defense Technology (NUDT), Changsha, China, in 2017, 2012 and 2010, respectively.
Now, he is an assistant research fellow in the National Laboratory for Parallel and Distributed Processing, National University of Defense Technology, P.R.China.
He worked as a research assistant in Chinese University of Hong Kong, from May 2014 to Oct. 2014.
His research interests include structured sparse learning, online and distributed optimization, and deep learning for graph and graphical models.
\end{IEEEbiography}

\begin{IEEEbiography}[{\includegraphics[width=1in,height=1.25in,clip,keepaspectratio]{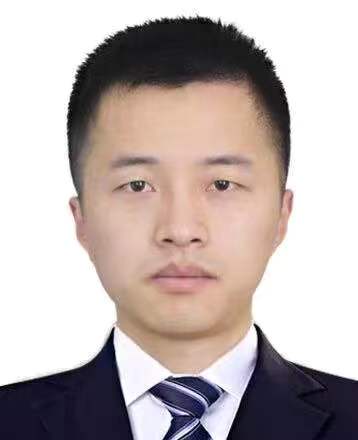}}]{Sen Yang} received his B.E. and M.E. degree from the National University of Defense Technology (NUDT), Changsha, China, in 2015 and 2017 respectively. He is currently doing a doctorate in computer science. His research interests include text generation and knowledge graph Q\&A.
\end{IEEEbiography}

\begin{IEEEbiography}[{\includegraphics[width=1in,height=1.25in,clip,keepaspectratio]{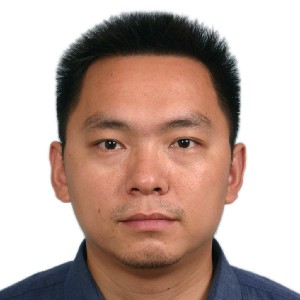}}]{Feng Liu}  received the Ph.D., M.S., and B.S. degrees in computer science and technology from the National University of Defense Technology (NUDT), Changsha, China, in 2006, 2002 and 1999, respectively.
Now, he is an associate research fellow in the National Laboratory for Parallel and Distributed Processing, National University of Defense Technology, P.R.China.
His research interests include distributed computing, big data.
\end{IEEEbiography}

\begin{IEEEbiography}[{\includegraphics[width=1in,height=1.25in,clip,keepaspectratio]{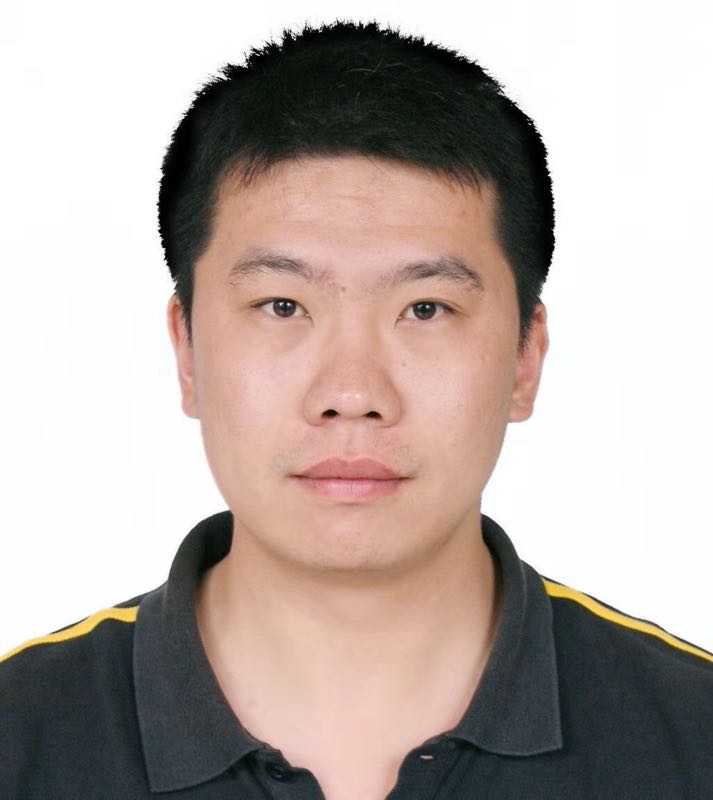}}]{Feng Huang} received the Ph.D., M.S., and B.S. degrees in computer science and technology from the National University of Defense Technology (NUDT), Changsha, China, in 2017, 2012 and 2010, respectively.
Now, he is an assistant research fellow in the National Laboratory for Parallel and Distributed Processing, National University of Defense Technology, P.R.China.
His research interests include datacenter network, cloud computing, and big data.
\end{IEEEbiography}

\EOD
\end{document}